\pdfoutput=1

\documentclass[11pt]{article}

\usepackage[]{ACL2023}

\usepackage{times}
\usepackage{latexsym}
\usepackage{xspace}
\usepackage{multirow}
\usepackage{graphicx}
\usepackage{booktabs}
\usepackage{ dsfont }
\usepackage{amsmath}
\usepackage{graphicx}
\usepackage{inconsolata}
\usepackage{algpseudocode}
\usepackage{algorithm}
\usepackage{caption}
\usepackage{subcaption}

\graphicspath{ {./images/} }

\usepackage{listings}
\usepackage[T1]{fontenc}

\usepackage[utf8]{inputenc}

\usepackage{microtype}

\usepackage{inconsolata}

\usepackage[capitalize]{cleveref}

\usepackage{soul} 

\newcommand\critic{$\operatorname{LM}_{\textrm{critique}}$\xspace}
\newcommand\repair{$\operatorname{LM}_{\textrm{refine}}$\xspace}
\newcommand\mcritic{\operatorname{LM}_{\textrm{critique}}}
\newcommand\task{$\operatorname{LM}_{\textrm{task}}$\xspace}
\newcommand\mtask{\operatorname{LM}_{\textrm{task}}}
\newcommand\objc{\textsc{critique}\xspace}
\newcommand\objp{\textsc{predict}\xspace}
\newcommand\objr{\textsc{refine}\xspace}
\newcommand\objd{\textsc{directrefine}\xspace}
\newcommand\supervised{\textsc{supervised}\xspace}

\title{RL4F: Generating Natural Language Feedback with Reinforcement Learning for Repairing Model Outputs}

\author{
Afra Feyza Akyürek$^{1}$ \And
Ekin Akyürek$^{2}$ \And
Aman Madaan$^{4}$ \And
Ashwin Kalyan$^{5}$ \AND
Peter Clark$^{5}$ \And
Derry Wijaya$^{1,3}$ \And
Niket Tandon$^{5}$\AND
\normalfont{$^1$Boston University} \quad \normalfont{$^2$MIT CSAIL}
\quad \normalfont{$^{3}$Monash University Indonesia}\\\normalfont{$^{4}$Language Technologies Institute, Carnegie Mellon University}\\\normalfont{$^{5}$Allen Institute for Artificial Intelligence}
}

\begin{document}
\maketitle
\begin{abstract}
Despite their unprecedented success, even the largest language models make mistakes.
Similar to how humans learn and improve using feedback, previous work proposed providing language models with \textit{natural language feedback} to guide them in repairing their outputs. Because human-generated critiques are expensive to obtain, researchers have devised learned critique generators in lieu of human critics while assuming one can train downstream models to utilize generated feedback. However, this approach does not apply to black-box or limited access models such as ChatGPT, as they cannot be fine-tuned. Moreover, in the era of large general-purpose language agents, fine-tuning is neither computationally nor spatially efficient as it results in multiple copies of the network. In this work, we introduce RL4F (Reinforcement Learning \textit{for} Feedback), a multi-agent collaborative framework where the critique generator is trained to maximize end-task performance of GPT-3, a fixed model more than 200 times its size. RL4F produces critiques that help GPT-3 revise its outputs. We study three datasets for action planning, summarization and alphabetization and show relative improvements up to 10\% in multiple text similarity metrics over other learned, retrieval-augmented or prompting-based critique generators.\footnote{A significant portion of this work was done while Feyza was an intern at Allen AI. Correspondences to Afra Feyza Akyürek (\texttt{akyurek@bu.edu}).}
\end{abstract}

\begin{figure}[t!]
    \centering
    \includegraphics[width=0.48\textwidth]{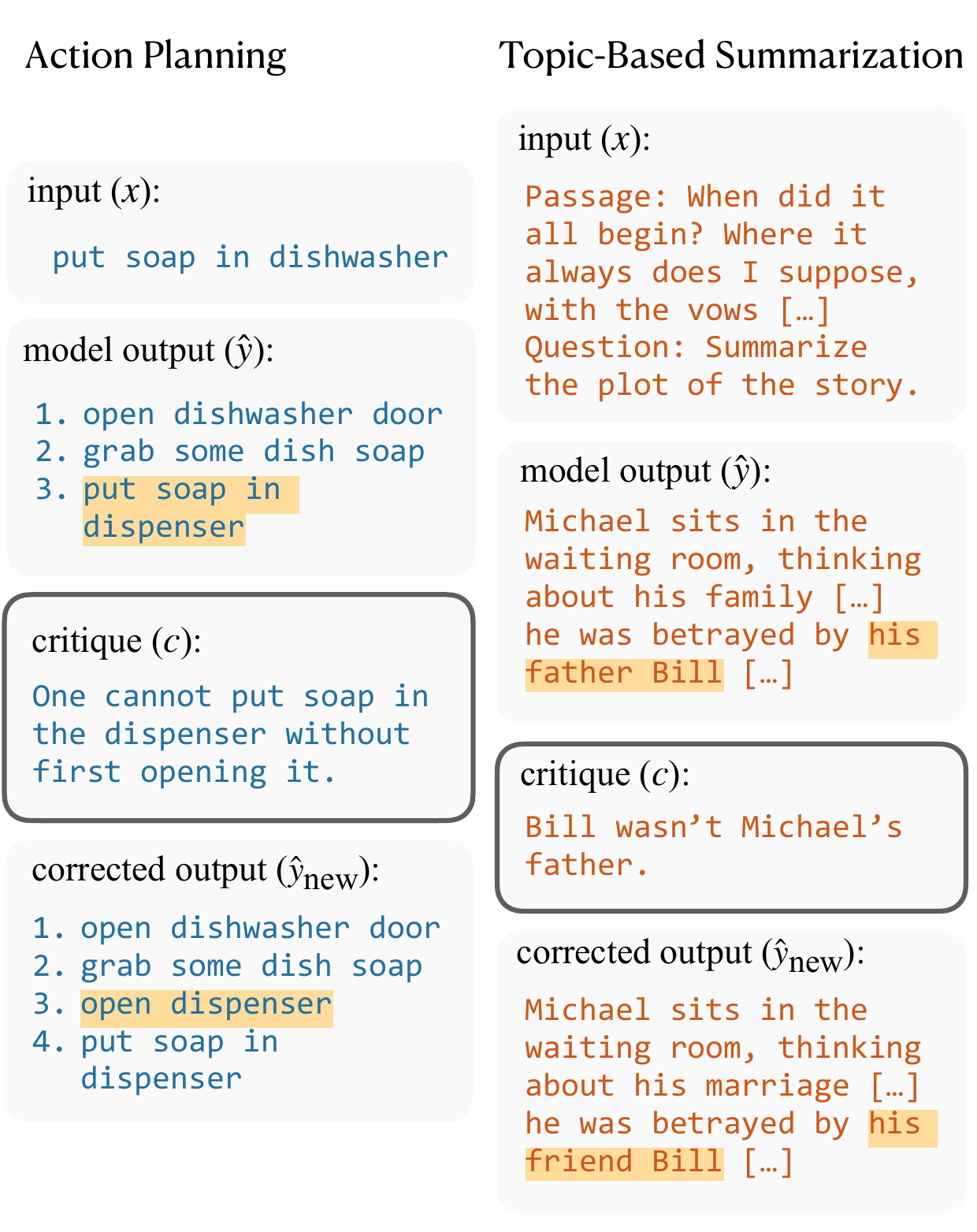}
    \captionsetup{font=footnotesize}
    \caption{Two examples for action planning \cite{tandon2021interscript} and summarization \cite{saunders2022self} tasks showcase a scenario where initial predictions by a learned model ($\hat{y}$) are incorrect. Human-written critiques ($c$) indicate errors in model outputs. While humans can reliably critique each other, machines lack such ability. This paper studies a multi-agent collaborative framework where one language model can generate critiques to improve its peer's performance.}
    \label{fig:teaser}
\end{figure}

\section{Introduction}

Correcting model outputs is a pressing challenge in natural language generation \cite{ribeiro-etal-2018-local, reid2022learning}, emerging across many use-cases such as style transfer \cite{mallinson-etal-2020-felix, malmi-etal-2022-text}, grammatical \cite{lichtarge-etal-2019-corpora} or factual error correction \cite{mitchell2022memory}, debiasing and detoxification \cite{schick-etal-2021-self}. Unlike humans who can understand natural language feedback and improve using the information, most of the previous work relied on sequence tagging \cite{reid2022learning}, retraining from scratch \cite{sun-etal-2019-mitigating} or parameter editing \cite{mitchell2022fast} to repair model predictions.

Recently, researchers show that large language models can correct their answer given more sophisticated feedback formulated in natural language \cite{schick2022peer,saunders2022self}. For example, in \cref{fig:teaser}, we present sample feedback for two tasks. Both of these examples exemplify the case where the initial model outputs $\hat{y}$ have flaws. In topic-based summarization, an automatically generated summary of a story involves factually incorrect statements such as \textit{``... he was betrayed by his father Bill ...''} where an appropriate critique is \textit{``Bill is not Michael's father''}. In action planning, given a goal $x$, the objective is to generate a set of steps $y$ to achieve the goal. The initial sequence of actions in \cref{fig:teaser}, denoted by $\hat{y}$, has a missing a step. The human-written natural language critiques $c$ describe the ways in which $\hat{y}$'s are incorrect and $\hat{y}_\text{new}$ denotes the corrected prediction conditioned on the critique. Note that in many situations helpful critiques do not necessarily reproduce an entire answer---they may simply point out one way in which the answer could be improved. 

Researchers use crowd-sourcing to collect critiques for model outputs \cite{saunders2022self}.
However, collecting feedback from humans is infeasible in an online setting where a model is required to produce a rapid stream of outputs. The goal of this paper is to shed light on whether the task of critiquing language model predictions can be effectively passed on to an external agent while keeping the language model itself intact.

Our multi-agent collaborative framework involves two language models where one model's job is to criticize the other as the latter performs a task of interest, such as summarization. This setting comprises a task model, denoted by \task, which learns the mapping from an input $x$ (e.g. passage) to a ground truth output $y$ (e.g. summary); and a critiquing model \critic which provides natural language critiques for \task's outputs $\hat{y} \sim \mtask(x)$. The framework can additionally involve a separate model (say \repair) for repairing model outputs conditioned on critiques. We follow past work \cite{schick2022peer}, and merge \task and \repair into a single model. Hence, in addition to predicting $y$ given $x$, \task is also tasked to improve its initial output conditioned on a critique $\hat{c}$ sampled from $\mcritic(x,\hat{y})$.

We introduce RL4F (Reinforcement Learning for Feedback Generation), a cascade \cite{dohan2022language} of two language models for automatic critique generation and refinement. RL4F is trained to maximize target task performance of \task by learning to provide critiques for its outputs via \critic. RL4F advances retrieval-based methods with learned critique generation \cite{madaan2022memory}. Unlike previous work which teaches \task to read a crowd-sourced set of critiques \cite{schick2022peer, saunders2022self}, RL4F learns the particular set of critiques that will steer \task into improving its predictions without requiring any updates to \task parameters. Treating \task as fixed is especially important in era of limited-access large language models which are costly, if not impossible, to fine-tune.

RL4F is illustrated in \cref{fig:method}(a,c). Previous work demonstrate that language models smaller than roughly 50 billion parameters lack the ability to understand and act upon a natural language critique \cite{saunders2022self, bai2022constitutional}. Therefore, we chose GPT-3 as the \task model which is a clear example of an inaccessible LM that shows this ability. While RL4F is general enough to accommodate an ensemble of feedback generators, in this work we focus one single model as \critic for simplicity.

In summary, this work\footnote{Code for the experiments is released under \url{https://github.com/feyzaakyurek/rl4f}.}: 
\begin{itemize}
    \item Presents a reinforced critique generator which advances simple supervision in improving the end-task performance without retraining the downstream model. 
    \item Demonstrates effectiveness of RL4F on three tasks: topic-based summarization, action planning and alphabetization (e.g. sorting a list of words alphabetically) with relative improvements up to 10\%.
    \item Showcases that RL4F exhibits promising scaling properties and remains to be useful when applied iteratively.
\end{itemize}

\section{Related Works}

Past works differ to a large extent with respect to what they call human feedback and how they make use of it. In this section, after elucidating the use of human feedback in previous works, we briefly describe connections of RL4F to the parameter-efficient fine-tuning and discrete prompt learning literature.

\subsection{What kind of feedback is used and where does it originate?}
Human feedback on model predictions come in different flavors. The most notable ones include (1) binary feedback, e.g. thumbs up/down and pairwise comparisons \cite{ouyang2022training, bai2022constitutional, gao2022simulating}, (2) natural language critiques \cite{tandon-etal-2022-learning, schick2022peer, madaan2022memory, saunders2022self, murty2022fixing, chen2023teaching, madaan2023self} and (3) direct textual refinements to outcomes \cite{scheurer2022training,shi2022life}.

\citet{bai2022constitutional} introduce what they call Reinforcement Learning from AI Feedback (RLAIF) in which they replace human preference labels with those of the model's itself; the model is prompted to evaluate its own predictions in consideration of human values and preferences. In a similar vein, \citet{gao2022simulating} use accuracy for extractive question answering as a reward signal when fine-tuning their policy model.

In another thread,\citet{schick2022peer} use comments from forums and Wikipedia edit histories as natural language feedback. \citet{scheurer2022training} and \citet{shi2022life} collect human natural language critiques and associated refinements. They then fine-tune the task model on the refinements. Our work is similar to these works in that we also use human-generated critiques in the first stage of our algorithm. Aside from human-written critiques, we additionally use synthetically generated critiques in the absence of the former.

\subsection{How is feedback used?}
An overwhelming majority of past work simply fine-tunes their task model using human feedback; whether it is a general purpose language model \cite{ouyang2022training, bai2022constitutional} or a task-specific model \cite{shi2022life, gao2022simulating, saunders2022self, scheurer2022training, schick2022peer}. \citet{tandon-etal-2022-learning} differently fine-tunes a separate corrector model which takes in a \textit{retrieved} critique utterance to correct initial outputs. Similarly, \citet{madaan2022memory} retrieves from a memory of previous critiques to improve GPT-3 predictions via few-shot prompting.

Our work separates from existing work by focusing on critique generation and harnessing critiques that yield better final outcomes by \task. Similar to \citet{schick2022peer}, we effectively propose a multi-agent setup by disentangling critique generation and conditional refinement. Differently, we keep the latter model intact and only train the critique generator \critic via reinforcement learning. Moreover, we take a step forward by leveraging end task data for the first time and directly optimize the critique generation process to improve final task performance. In contrast to RLHF whose policy network (\task) is trained to maximize human alignment \cite{wiener1960some}, our policy network (\critic) is trained to bootstrap end-task success of \task. Our proposal RL4F is orthogonal to RLHF; in fact we use an RLHF fine-tuned checkpoint in our experiments. For further discussion, please refer to \citet{fernandes2023bridging} who catalogue different approaches on integrating natural language feedback to textual generations.

\subsection{Adapters \& Discrete Prompt Learning}

A large body of existing work finds that parameter-efficient fine-tuning, often referred to as adapters \cite{pfeiffer2020AdapterHub} is as effective as full fine-tuning while being computationally cheap. RL4F can also be interpreted as an alternative ``adapter'' under the strict setting where only textual access to task model is available. Furthermore, our work can also be viewed from the perspective of learning discrete prompts for language models. Past work propose to generate knowledge pieces \cite{liu2022rainier} or arbitrary textual snippets \cite{deng2022rlprompt} which they append to the input via reinforcement learning. These works are different than ours in that their policy is conditioned solely on the input $x$ whereas in our case we sample critiques of machine-generated predictions based on $x$ \textit{and} $\hat{y}$.

\begin{figure*}
    \centering
    \includegraphics[width=0.97\textwidth]{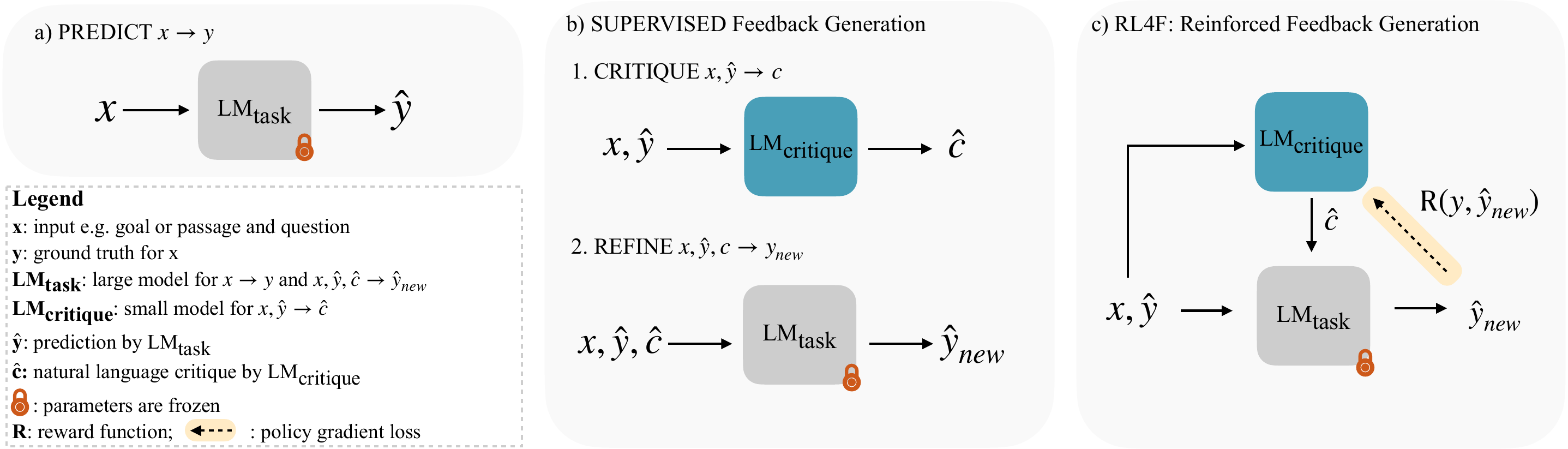}
    \captionsetup{font=footnotesize}
    \caption{a) A downstream task model takes in an input (e.g. a passage and a question) and predicts the output (e.g. summary). b) Past work proposed using a supervised learning scheme \cite{saunders2022self, schick2022peer} or retrieval \cite{madaan2022memory} for critique generation (\objc) and refinement tasks (\objr). In our setting, we only train \critic and parameters of the task model are left unchanged. c) RL4F uses \critic that was produced as a result of the training in part b. Using task data pairs (e.g. passages and summaries) we continue fine-tuning \critic with policy gradient such that critiques steer \task to produce better outputs.}
    \label{fig:method}
\end{figure*}

\section{Background}
\label{sec:back}

The problem of learning from critiques entails two major challenges: (1) generating critiques and (2) the task of refining initial answers based on a critique. In our experiments (\cref{sec:exps}), GPT-3 responds very well to ground-truth critiques. This observation suggests that given quality critique, GPT-3 is indeed able to improve a potentially erroneous prediction for the better. Hence, in this study we focus our efforts on (1). Our ultimate goal is to reach, and eventually exceed, human-level critique performance using machines.

Following \citet{saunders2022self}, we identify four primary functions towards studying the problem of learning with natural language critiques. First is \textsc{predict}: the base task of predicting without using critiques to model $x \to y$. As an example, if $x$ is a passage, $y$ is the summary (see \cref{fig:teaser}). Moreover, we refer the task of learning to generate critiques $x,\hat{y} \to c$ where $\hat{y} \sim$ $\mtask(x)$ as \textsc{critique}. Lastly, we call the conditional refinement objective $x,\hat{y},c \to y$ as \textsc{refine} and repairing an answer without a critique $x,\hat{y} \to y$ as \textsc{directrefine}\footnote{Additionally, \citet{saunders2022self} put forth the tasks of \textit{critiqueability} to gauge whether an output requires a critique in the first place and \textit{helpfulness} referring to if a sampled critique is useful. We let the critique generator model to implicitly take care of these tasks.}. We use $\hat{y}$ and $\hat{c}$ notation to indicate an estimate of ground truth $y$, and similarly for $c$, from a respective model.

\subsection{\textsc{supervised}: Supervised Learning for Critique Generation}
\label{subsec:supervised}
We initialize \critic to be a pretrained encoder-decoder-type model and fine-tune it to generate critiques satisfying the \objc objective $x,\hat{y} \to c$ using natural language critiques. Namely, if \critic is parameterized by $\theta$ we maximize $\mathds{E}\left[\log p_\theta(c|x,\hat{y})\right]$. We delegate \objp and \objr tasks to GPT-3 via in-context learning. The procedure is depicted in \cref{fig:method}a-b.

The main difference of our implementation of \textsc{supervised} to that of \citet{saunders2022self}'s is that we rely on separate models for \objc and the rest of the tasks while they train a single GPT-3-style model to collectively achieve \objp, \objc, \objr and \objd; effectively merging \critic and \task into a single model. While this may seem parameter-efficient, our version has a few key advantages. First, leaving any \task model intact (parameters frozen) enables us to work with models that are already-deployed as \task and those with expensive training and inference processes. Moreover, our approach refrains from disturbing overall integrity of a general-purpose language model by conditioning it to a specific task. Lastly, training \critic, which is multiple orders of magnitude smaller than GPT-3, is much more computationally efficient and therefore accessible to a broader range of users.

\subsection{Direct-Refinement}
\label{sec:self-refinement}
\citet{madaan2023self,chen2023teaching} propose that using the critiques from the model itself via few-shot prompting results in improved performance. On the contrary, \citet{saunders2022self} and \citet{bai2022constitutional} argue that direct refinement (as denoted with \objd in this work) i.e. the practice of prompting a language model without self-generated critiques to \textit{directly} repair its own answers  proves a stronger baseline, especially when the model size is large >50B. They hypothesize that this is primarily due to model's initial answers getting increasingly more difficult to self-critique as the model size grows. In fact, both \citet{saunders2022self} and \citet{bai2022constitutional} showed that their largest model achieves superior end-task performance when performing \objd than refining using self-generated critiques. Hence, we use Direct-Refinement as a baseline and describe how we implement it via in-context learning in \cref{sec:exps} while providing further discussions in \cref{app:self-refine}.

\section{RL4F: Reinforcement Learning for Feedback Generation}

\supervised is straightforward to implement but it does not make use of any final task data ($x \to y$) that is usually more abundant than natural language critiques. Moreover, it fails to provide ground for adaptation when the critiques in the train set are generally applicable but not entirely tailored to improving a target model. We describe RL4F where we follow supervised training with policy gradient learning using end-task data in order to generate critiques. We assume that the task model \task is already deployed and treat it as a fixed module. In all of our implementations we train the natural language critique generator \critic alone. In both \supervised and RL4F, \critic takes in the input $x$ and an initial prediction $\hat{y}$ and produces a (natural language) critique $\hat{c}$: 
\begin{equation}
\label{eq:sampled_critique}
    \mcritic(x,\hat{y}) = \hat{c} ~.
\end{equation}

\cref{fig:method}c provides an illustration of RL4F. We implement \task as GPT-3 given its adaptability into new tasks using few-shot prompting. Our implementation which is primarily based on the RL4LMs library\footnote{\url{https://github.com/allenai/RL4LMs}} \cite{ramamurthy2022reinforcement} will be publicly available.

\paragraph{Learning via Policy Gradient} 
We warm-start RL4F by first fine-tuning \critic for \objc which we defined as is the supervised objective of learning to generate natural language critiques $c$ conditioned on $x,\hat{y}$. We continue fine-tuning the policy network (\critic) to maximize the reward using Proximal Policy Optimization \cite{schulman2017proximal}. We utilize the implementation of PPO provided by \citet{ramamurthy2022reinforcement} and refer the readers to the original work about the details for KL-regularized PPO objective. While any policy gradient approach could be used e.g. REINFORCE \cite{williams1992simple}, our initial experiments showed that PPO works best in this setting.

Pseudocode for RL4F is provided in \cref{algo:rl4f} where we use two sets of in-context examples for prompting GPT-3. We define $E$ to be a set of in-context-learning examples in the form of $(x,y)$ to get GPT-3 solve \objp. Similarly, $E^c$ contains in-context examples to prompt GPT-3 to fix an initial attempt $\hat{y}$ into $y$ conditioned on the natural language critique $c$ which we termed as \objr; $E^c = \{(x,\hat{y},c,y)\}$. As per our reward function, we opt to use a lexical similarity metric ROUGE (1/2/L) \cite{lin2004rouge} in \cref{eq:rew1} for planning and summarization datasets. Measuring ROUGE is computationally fast, making it easy to use in an online learning setting. Reward is only collected at a terminal stage i.e. either when the end of sentence token is produced or maximum number of tokens is reached. 

\begin{equation}
\textrm{R}(\hat{y}, y) = \operatorname{mean} \left(\textrm{R1}(\hat{y},y),\textrm{R2}(\hat{y},y), \textrm{RL}(\hat{y},y)\right) \label{eq:rew1}
\end{equation}
\begin{algorithm}[t]
\footnotesize
\begin{algorithmic}
  \State {\textbf{Input:}} Dataset $\mathcal{D}=\{(x^{i}, \hat{y}^i, y^{i})\}_{i=1}^{N}$ of size $N$
  \State {\textbf{Input:}} Initial \critic, \task
  \State {\textbf{Input:}} In-context examples for refinement $E^c$
  \Repeat
      \State Sample mini-batch $\mathcal{D}_m=\{(x^{m}, y^{m}, \hat{y}^{m})\}_{m=1}^{M} \sim D$
      \State Sample $\hat{c}\sim \operatorname{LM}_{\textrm{critique}}(x,\hat{y})$ for $D^m$ in parallel
      \State Sample $\hat{y}_{new} \sim \operatorname{LM}_{\textrm{task}}(E^c,x,\hat{y},\hat{c})$ for $D^m$
      \State Compute KL-regularized rewards $R_t$ \Comment{\cref{eq:rew1}}
      \State Compute the advantage estimate $\hat{A}_t$
      \State Update the \critic by maximizing the PPO objective
  \Until{convergence} and 
  \Return \critic
\end{algorithmic}
\caption{RL4F \newline {\footnotesize Pseudocode of the algorithm used to train feedback model.}}
\label{algo:rl4f}
\end{algorithm}
\section{Datasets}

\subsection{Topic-Based Summarization}
\citet{saunders2022self} crowd-sourced natural language critiques for topic-based summarization. The train, validation and test sets contain 14230, 1150 and 2658 tuples of $(x,\hat{y},\hat{c})$. The dataset provides multiple questions for a given passage each inquiring about a different aspect. Given a passage and question ($x$) multiple summaries are sampled from the model. Human annotators provide natural language critiques for the answers along with improved summaries. One example is provided in \cref{fig:teaser} and more are available in the Appendix (\cref{tb:pred_samples_summ}).

\subsection{Interscript}
Interscript \cite{tandon2021interscript} is an action planning dataset for everyday tasks such as ``put on a costume'' or ``play musical chairs''. Each goal $x$ is associated with a sequence of ground truth actions $y$. Along with $x,y$ pairs, it contains erroneous action plans $\hat{y}$ and natural language critiques $\hat{c}$ suggesting a fix. An example is provided in \cref{fig:teaser} for ``put soap in dishwasher''. Other examples of critiques are ``You need to have music to play musical chairs.'' and ``You need to pay for the costume before leaving the store''. More examples are available in the Appendix (see \cref{tb:pred_samples_plan}). Interscript represents a low-resource scenario: it contains 253, 45 and 169 examples for train, validation and test sets where each example contains 1-4 reference texts.

\subsection{Synthetic Task: Alphabetization}
\label{sec:datasets_alph}
We synthetically generate a task for alphabetically sorting a list of words with lengths ranging between 3-12. We use the lexicon \#11 by \citet{bigenglishwordlist} which contains 43K unique English words. Given an unsorted list and a ground truth sorting of the list we identify 5 operations to sample a incorrect sorting of $y$ denoted by $\hat{y}$ and associated critique $c$ articulating what is wrong about $\hat{y}$ in natural language. One example is shown below:
\begin{lstlisting}
    x: mug, greek, book, house
    y: book, greek, house, mug
    $\hat{y}$: book, greek, house
    c: The word mug is missing.
\end{lstlisting} 

The operations we use for distortion are \texttt{REORDER}, \texttt{REPLACE}, \texttt{ADD}, \texttt{REPEAT} and \texttt{REMOVE} (shown above). We also leave majority of sorted lists intact for which the ground truth critique is ``The list is correctly sorted''. We use a total of 40K examples for warm-starting \critic for the \objc objective and another 10K, 1K and 1K examples for PPO stage, for train, dev and test splits, respectively. Examples delineating other operations in action and corresponding natural language critiques are provided in \cref{sec:dataset_proc}.

In alphabetization, we use Inverse-Levenstein distance for the reward function $\text{R}$ as defined in \cref{eq:ilev} where $\lvert\cdot\rvert$ measures length of the list. Levenstein distance is a form of edit distance for single character edit operations such as removal, insertion and substitution in a string. We count word-level operations rather than character-level. Note that the higher inverse-Levenstein score of a predicted ordering, the closer it is to the alphabetically sorted version. The sorted list gets the maximum reward of 1.

\begin{equation}
\label{eq:ilev}
R(\hat{y}, y) = 1 - \frac{\text{Levenstein}(\hat{y}, y)}{\max(\lvert\hat{y}\rvert, \lvert y \rvert)}
\end{equation}

\begin{table*}
\centering

\resizebox{0.97\textwidth}{!}{%
\begin{tabular}{lccc|ccc}
\toprule
\multicolumn{1}{l}{} & \multicolumn{3}{c}{Interscript}  & \multicolumn{3}{c}{Topic-Based Summarization}\\
\textbf{Source of Critiques} & \textbf{BLEURT \textuparrow} & \textbf{BERTScore\textuparrow} & \textbf{R1/R2/RL\textuparrow} & \textbf{BLEURT\textuparrow} & \textbf{BERTScore\textuparrow} & \textbf{R1/R2/RL\textuparrow}\\ \midrule
Direct-Refinement  & -1.07  & 86.97  & 15.8 / 0.9 / 15.5 & 0.09 & 93.1 & 54.3/46.0/50.9 \\
\supervised               & -1.02  & 86.99  & 19.4 / 0.5 / 18.5 & 0.06 & 92.9  & 53.2/46.4/50.7 \\
MemPrompt & -1.18 & \textbf{87.45} & 16.9 / \textbf{1.9} / 16.7 & 0.09 & 91.9 & 48.8 /40.4/45.6 \\ 
RL4F (Ours)              & \textbf{-0.92} & 87.23 & \textbf{22.1} / 0.9 / \textbf{21.3}    & \textbf{0.10} & \textbf{93.6} & \textbf{55.1}/\textbf{48.2}/\textbf{52.6}\\\midrule
\textit{With gold feedback}  & \textit{-0.69} & \textit{89.56} & \textit{40.7} / \textit{6.8} / \textit{39.1}   & \textit{0.22}  &  \textit{94.2} & \textit{58.3}/\textit{50.3}/\textit{55.8}           \\ \bottomrule
\end{tabular}%
}
\captionsetup{font=footnotesize}
\caption{Results for action sequence generation with Interscript \cite{tandon2021interscript} and topic-based summarization by \citet{saunders2022self}. We evaluate the performance of different sources for natural language critiques in steering \task to improve its predictions. Best scores in each column are made bold. 
We compare our method, RL4F, to three strong baselines and human-generated critiques. Self-Refinement prompts GPT-3 to self-repair its answer. 
MemPrompt uses memory to store human-generated critiques to previous outputs \cite{madaan2022memory}.  ROUGE and BERTScore are out of 100 while BLEURT can be negative or positive and should be used in comparing different methods.
}
\label{tb:main_res}
\end{table*}

\begin{table}[t]
\centering
\resizebox{0.5\textwidth}{!}{%
\begin{tabular}{lrr}
\toprule
\textbf{Source of Critiques} & \multicolumn{1}{l}{\textbf{Exact Match}} & \multicolumn{1}{l}{\textbf{Inverse Levenstein}} \\ \midrule
Initial Outputs ($\hat{y}$) & 63.7 & 0.91 \\ \midrule
Fine-tuning \texttt{davinci} & 55.3 & 0.89 \\
MemPrompt & 57.8 & 0.89\\
Direct-Refinement        &            65.9                   &                    \textbf{0.92}            \\
\supervised                     &              38.9               &                0.82              \\
RL4F               &         \textbf{66.1}           &             \textbf{0.92}            \\\midrule
\textit{With gold feedback}  &        75.9              &      0.94                   \\ \bottomrule
\end{tabular}%
}
\captionsetup{font=footnotesize}
\caption{Results for alphabetization. Best scores are highlighted. Initial Outputs are obtained from GPT-3 (\texttt{code-davinci-002}) via in-context learning. \supervised critiques misguides GPT-3, hurting its initial performance, as with MemPrompt. RL4F improves over the performance of \supervised model by 27 absolute points. Self-Refinement around the same as RL4F. In \cref{fig:iterative_alph}, we further discuss advantages of RL4F over Self-Refinement when we sample and refine iteratively.}
\label{tb:alph}
\end{table}

\section{Experiments and Results} 
\label{sec:exps}

Our experiments are designed to test effectiveness of RL4F, along with other sources of critiques, in both natural and controlled settings. In our evaluations, we test the usefulness of critiques by looking at the final task performance rather than evaluating generated critiques themselves; as multiple critiques may lead to the same correct answer.

\paragraph{Sampling Critiques} We sample critiques from \critic as in \cref{eq:sampled_critique} by first concatenating the input and initial prediction. The specific input format for \critic we use for Interscript is given below and the other two can be found in \cref{sec:app_exp}. We initialize \critic with pretrained T5-large which is a 0.77M parameter encoder-decoder type language model trained on large web text \cite{raffel2020exploring}.

\begin{verbatim}
    Goal: {goal}
    Steps: {initial_answer}
\end{verbatim}

\paragraph{Downstream Predictions} 

\begin{figure}
    \centering
    \includegraphics[width=0.45\textwidth]{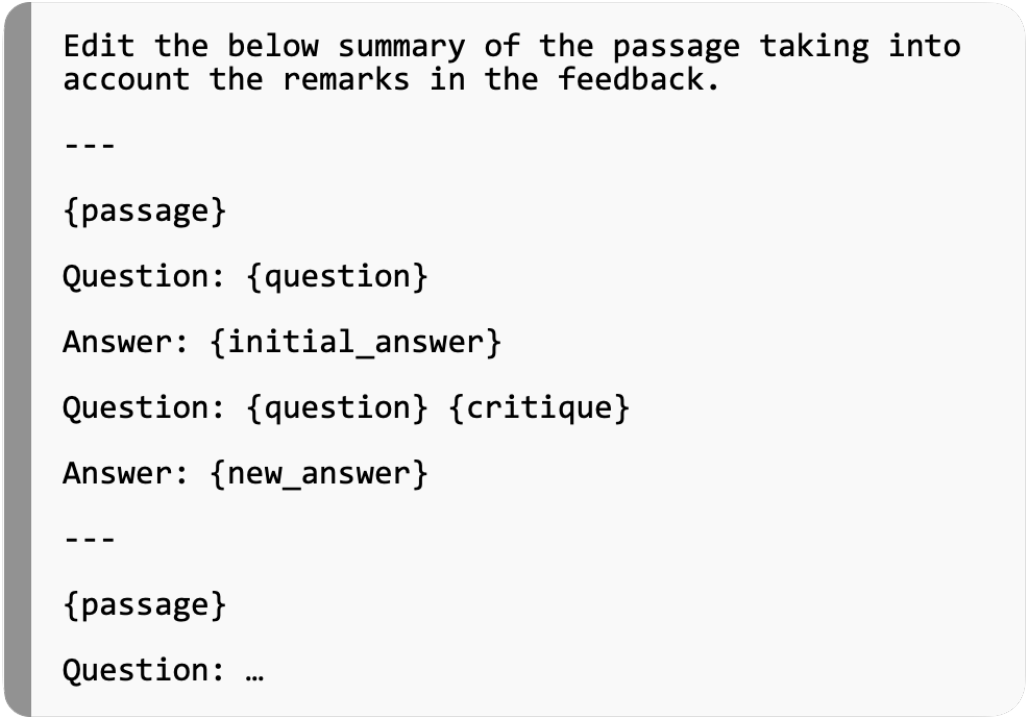}
    \captionsetup{font=footnotesize}
    \caption{Prompt template for topic-based summarization. We ask GPT-3 to refine the initial prediction by using \texttt{critique}.}
    \label{fig:summ_prompt}
\end{figure}

In our experiments, we consider GPT-3 as the \task model. GPT-3 can handle a wide range of tasks with prompting---using a handful of task examples in the input and without requiring task-specific fine-tuning \cite{brown2020language}. GPT-3 is not only able to tackle numerous tasks conveniently but also can refine initial predictions when given a natural language critique \cite{madaan2022memory}. Since our setting requires \task model to be able to model both the main task objective $x \to y$ and the refinement objective $x,\hat{y},c \to y$, GPT-3 is a suitable candidate that can adapt to both, using few-shot exemplars. The prompt template we use for the latter is shown in \cref{fig:summ_prompt} where we provide the model with an initial attempt to the question \texttt{initial\_answer} and re-sample a revised prediction conditioned on the \texttt{question} and \texttt{critique} for the summarization task. We use \texttt{code-davinci-002} checkpoint via OpenAI API\footnote{We find that only the largest 175B parameter GPT-3 can handle \objr.} and 3, 1 and 6 hand-written in-context examples for planning, summarization and alphabetizations tasks, respectively as we exhaust the 4096 token input limit.

In action planning, instead of resampling entire plans, we prompt GPT-3 to produce an edit operation on the initial plan $\hat{y}$. The set of edit operations identified in the original dataset are \textit{Insert}, \textit{Remove} and \textit{Reorder} where each critique comes with a corresponding edit operation. Note that, these operations can algorithmically be applied to $\hat{y}$. While \textit{Reorder} and \textit{Remove} are expected to refer to existing steps in $\hat{y}$, we expect \textit{Insert} to introduce a novel action. Hence, we stick with a generic lexical similarity metric in calculating reward (\cref{eq:rew1}) for this task. In summarization, we compare human-written summaries with the repaired model summaries. 

\paragraph{Baselines} We compare effectiveness of RL4F to \supervised which is described in \cref{subsec:supervised}. This is the closest baseline to the approach by \citet{saunders2022self} and \citet{schick2022peer} that abides by our condition that \task should remain unchanged. We use the same set of initial predictions $\hat{y}$ when comparing different critique generators.

In addition to \textsc{supervised}, we use a simple Direct-Refinement baseline where we ask \task to revise the initial prediction given a fixed critique ``Improve the answer.'' (\objd). The prompt template is otherwise the same as in other methods. We configure our in-context examples to show that not all $\hat{y}$ need to be repaired. Hence, \task is free to update the prediction or leave it as is when it is correct. Despite its simplicity, Direct-Refinement has been established as a strong baseline \cite{saunders2022self}.

Moreover, we compare to MemPrompt \cite{madaan2022memory}. In their work, authors study a setup where \task generates an \textit{understanding} along with the target output. For example, given a question \textit{``What sounds like good?''}, the model generates an understanding of the question \textit{``The question is asking for homonym.''} before saying \textit{``wood''}. In their critique retriever, they train a mapping to model $x$ into an understanding $u$. However \textit{understanding} is redundant in particular tasks e.g. summarization where the question is no different than $u$, thus throughout our experiments, we replace the learned retriever in MemPrompt with BM25 \cite{harter1975probabilistic}.

Lastly, we use human-written critiques (gold feedback) for \objr in getting \task to repair outputs and report this as an upperbound.

\subsection{Planning and Summarization}
Our main results for Interscript and topic-based summarization are provided in \cref{tb:main_res}. Given the free-from nature of the outputs, we evaluate planning and summarization tasks using text similarity metrics to capture  semantic and lexical similarities. We utilize learned metrics such as BLEURT \cite{sellam-etal-2020-bleurt} and BERTScore \cite{bert-score} along with ROUGE \cite{lin2004rouge}. We compare the performance achieved by using different sources of critiques to that of human-written critiques. Across all metrics, RL4F yields one of the closest outcomes to human-written critiques.\footnote{We have identified a handful of examples where a pair of train and test examples differs by only a single concept e.g. all occurrences of ``noodle'' in the train sample was replaced with ``food'' to produce the test sample. The goal and steps are the same otherwise. MemPrompt does exceedingly well on these 7 cases, hence performing occasionally higher, yet fails in the rest of the test/val examples.}

\subsection{Alphabetization}
We initialize our \critic using the synthetic critiques as described \cref{sec:datasets_alph}. Our results are provided in \cref{tb:alph}. For alphabetization we compute exact match and inverse-Levenstein scores as defined in \cref{eq:ilev}. As an additional baseline, we fine-tune \texttt{davinci} \cite{brown2020language} on the same train set as our RL4F.

Because of the synthetic procedure to create $x,\hat{y},c$ triplets, the generated $\hat{y}$ as well as $c$ do not necessarily reflect the kinds of errors that \task would do. We observe this in the scores of \supervised which fails to improve upon initial outputs. Nevertheless, RL4F procedure helps the policy network to capture a useful distribution of critiques, improving over \supervised by more than 27 absolute points. In this simple task, Direct-Refinement prompt also yields a competitive performance. Compared to full fine-tuning, we observe that despite training substantially fewer parameters RL4F achieves a significantly better accuracy. For a comparison to concurrent work \citet{madaan2023self}, please refer to the appendix.

\section{Analysis}

\paragraph{Scaling Properties}
\begin{figure}
    \centering
    \includegraphics[width=0.43\textwidth]{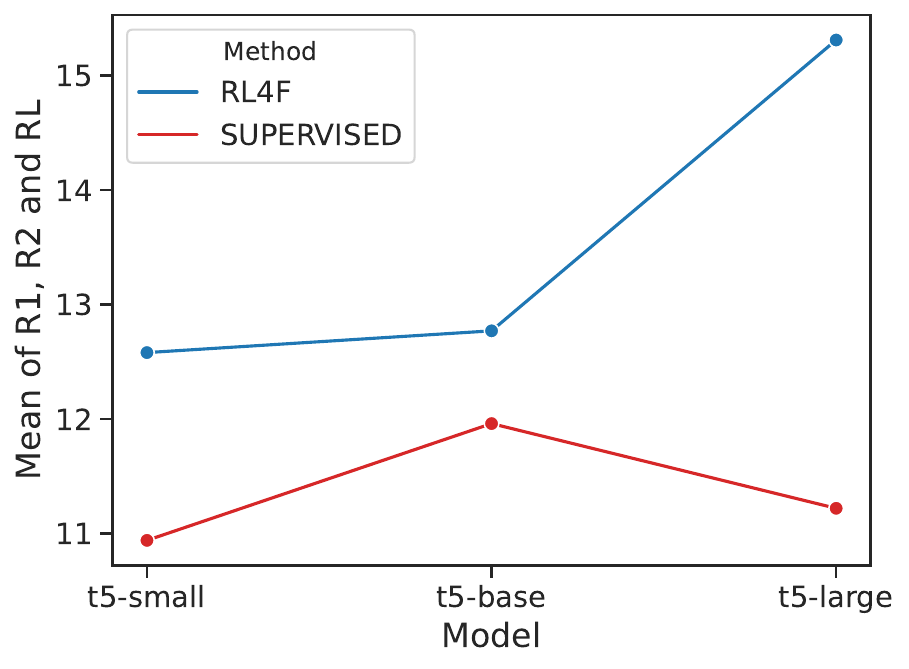}
    \captionsetup{font=footnotesize}
    \caption{Scaling properties of \supervised and RL4F on Interscript. We observe that RL4F greatly benefits from an increase in the number of parameters.}
    \label{fig:scale_interscript}
\end{figure}

While we use T5-large as our main model for all of our experiments to initialize \critic, we inquire about different model sizes. In \cref{fig:scale_interscript}, we consider three different model sizes to tackle Interscript ranging from 60M to 770M parameters. On the y-axis we provide average of three ROUGE scores for the generated plans. RL4F greatly benefits from an increase in the model size where a similar trend in \supervised is non-existent.

\paragraph{Semantic Drift} In goal-oriented training, semantic drift occurs when the strings produced by the policy begin diverging from initial language \cite{lee2019countering, blank1999new}. Although, RL4F does not guarantee that $\hat{c} \sim$ \critic will be natural language, we find minimal sign of semantic drift in the sampled critiques with respect to fluency and naturalness. In most cases, generated critiques are intelligible. We speculate that may be due to GPT-3 responding to natural language best than gibberish, though future work should look closely into this to make a more conclusive argument. We provide sample predictions from both models in \cref{sec:sample_predictions} for all three tasks.

\paragraph{Iterative Improvement}
\begin{figure}
    \centering
    \includegraphics[width=0.43\textwidth]{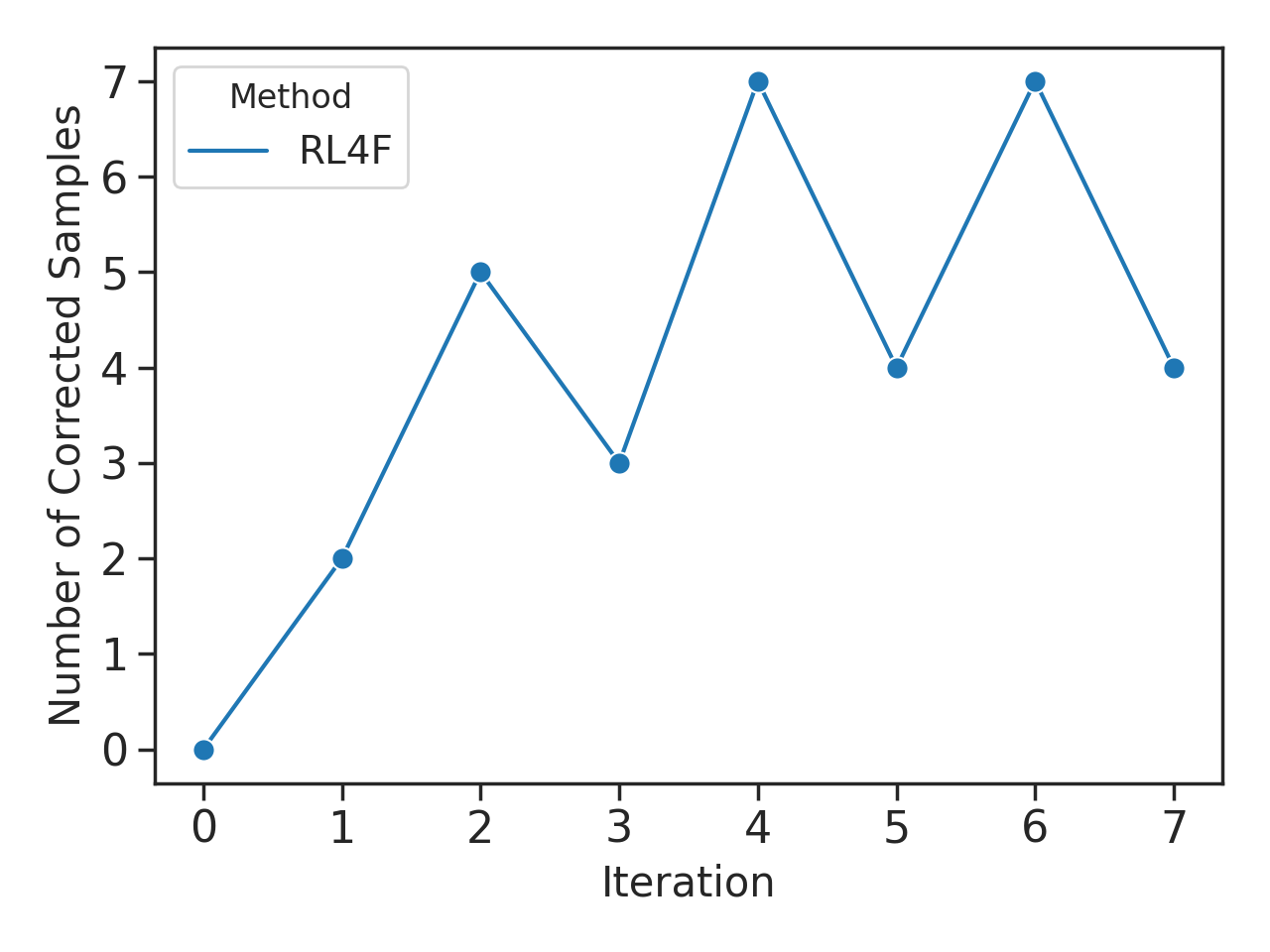}
    \captionsetup{font=footnotesize}
    \caption{We apply \textsc{refine} multiple times given critiques from \critic (RL4F) on alphabetization task. RL4F leads to a handful of more corrections when used iteratively.}
    \label{fig:iterative_alph}
\end{figure}

In \cref{sec:exps}, we provide results with applying only one round of critiques in alphabetization. Past work advocated for iterative editing \cite{reid2022learning, faltings-etal-2021-text} as opposed to one-shot editing. In \cref{fig:iterative_alph}, we sample and apply critiques from \critic to $\hat{y}$'s iteratively and see if the number of correctly sorted lists increase or decrease. Note that critiques may also lead to deteriorating performance as we are not eliminating the solved examples \textit{and} it is at \critic's discretion to declare a solution as correct e.g. by saying \textit{``This list is correctly sorted.''}. In fact, when we ask GPT-3 to simply improve its predictions iteratively (via Direct-Refinement as described in \cref{sec:self-refinement}), it occasionally scrambles an already correct ordering while not scoring any new points. In contrast, RL4F leads to up to 7 more corrections (see \cref{fig:iterative_alph}).

\section{Conclusion}

We have described a collaborative framework involving two language models where one model, the critique generator, is trained to improve the performance of the task model. We train the former via policy gradient learning while treating the task model as a black-box. We show that RL4F leads to superior final performance across three domains compared to other strong baselines without resulting as the critiques remain fluent and natural. Future work might focus on generalizing the critique generator into a mixture of experts allowing humans and other models to contribute to critiqueing procedure.

\section{Limitations}
RL4F is primarily targeted at improving final performance. While we have found that the critiques learned by RL4F remain natural, we do not introduce any explicit restraints preventing semantic drift. As though it may raise end-task performance, semantic drift would also hinder interpretability. Future work might study datasets that are not covered by this dataset and quantify semantic drift along with proposing measures to prevent it, as necessary. Moreover, this work does not provide an explicit mechanism to incorporate new critique labels that might become available in the future nor it identifies a framework that could combine critiques from multiple experts such humans and other machines. Lastly, we limit our analysis to GPT-3 and focus on a scenario where it is inefficient or impossible to train the task model while this may be a conservative assumption for other settings.

\section*{Acknowledgement}
We thank Anna Ivanova, Jacob Andreas, Zilu Tang, Shashank Gupta and Ashish Sabharwal for their valuable feedback on earlier drafts of this paper. Finally, we thank Rajkumar Ramamurthy and Prithviraj Ammanabrolu for helpful discussions on using their RL4LMs repository which facilitated the experiments of this work.

Afra Feyza Aky\"urek is supported in part by the U.S. NSF grant 1838193 and DARPA HR001118S0044 (the LwLL program). The U.S. Government is authorized to reproduce and distribute reprints for Governmental purposes. The views and conclusions contained in this publication are those of the authors and should not be interpreted as representing official policies or endorsements of DARPA and the U.S. Government. At MIT, Ekin Aky\"urek is supported by an MIT-Amazon ScienceHub fellowship and by the MIT-IBM Watson AI Lab. 



\bibliography{anthology,custom}
\bibliographystyle{acl_natbib}

\appendix

\section{Dataset Processing}
\label{sec:dataset_proc}
\paragraph{Action Planning}
Interscript is larger but we are only using a subset, removing distractors. The scripts used for data cleaning will be released along with the codebase.

\paragraph{Alphabetization} We sample initial predictions from GPT-3 for alphabetization some of which comprise multiple distortions simultaneously, yet we use one-step distortions to warm-start \critic. 

Given the following a pair of unsorted and sorted word lists e.g.
\begin{lstlisting}
    x: mug, greek, book, house
    y: book, greek, house, mug
\end{lstlisting}
below are the operations we used to create our data:
\begin{lstlisting}
    REORDER
    $\hat{y}$: book, house, greek, mug
    c: The word greek is placed in an
    incorrect position.
    
    REPLACE
    $\hat{y}$: book, greek, house, mud
    c: The word mug is replaced with mud

    REMOVE
    $\hat{y}$: book, greek, mug
    c: The word house is missing

    REPEAT
    $\hat{y}$: book, house, greek, house, mug
    c: The word house is repeated

    ADD
    $\hat{y}$: book, hair, greek, house, mug
    c: The word hair is not in the
    original list

    NOTHING
    $\hat{y}$: book, greek, house, mug
    c: The list is correctly sorted.
\end{lstlisting}

\section{Experiment Details}
\label{sec:appendix}

We use \texttt{code-davinci-002} as GPT-3 unless otherwise specified. We compute ROUGE implementation in the \texttt{datasets} library and set \texttt{use\_stemmer=True} for summarization and Interscript.

\subsection{Data Formats}
\label{sec:app_exp}
We use the following input formats for \critic:
\begin{itemize}
\item \textbf{Summarization}:
\begin{verbatim}
{passage}
Question: {question}
Answer: {initial_answer}
\end{verbatim}
\item \textbf{Planning}: 
\begin{verbatim}
Goal: {goal}    Steps: {steps}
\end{verbatim}
\item \textbf{Alphabetization}: 
\begin{verbatim}
{unsorted_list} ||| {initial_answer}
\end{verbatim}
\end{itemize}

We train separate models for each of the datasets and evaluate individually. We use T5-large provided by \texttt{transformers} library.

\subsection{Prompts for GPT-3}

We provide prompt templates used for Alphabetization and Interscript when prompting GPT-3 for \objr. Template for summarization is provided in the main text.
\begin{itemize}
    \item \textbf{Interscript:}
 \begin{verbatim}
Goal: {goal}
Steps: {steps}
Feedback: {critique}
Edit:
\end{verbatim}
    \item \textbf{Alphabetization:}
\begin{verbatim}
{unsorted_list} ||| {initial_answer}
Feedback: {critique}
Edit:
\end{verbatim}
\end{itemize}

In Direct-Refinement, templates remain the same and \texttt{critique}'s are replaced with ``Improve the answer.''. Exact prompt exemplars will be made available in the released code repository.

\subsection{Standard Deviations}
\label{app:std}
Standard deviations of R1/R2/RL scores across 5 runs in Interscript are 1.4/0.1/1.1 for \supervised and 0.5/0.4/0.5 for RL4F. 

\subsection{Hyperparameters}
In all of our experiments we use temperature 0 for prompting GPT-3 except when sampling initial predictions for alphabetization we set it to 0.5. We provide hyperparameters for RL4LMs \cite{ramamurthy2022reinforcement} in \cref{tbl:alph_hyperparams}, \cref{tbl:plan_hyperparams} and \cref{tbl:summ_hyperparams}.

\begin{table}
\centering
\footnotesize
\resizebox{0.5\textwidth}{!}{
\begin{tabular}{ll}
\toprule
\textbf{Model Params}
& \multicolumn{1}{c}{\textbf{value}} \\ 
\cmidrule{1-2}
supervised & batch size: $8$\\
& epochs: $10$ \\
& learning rate: $0.00001$ \\
& learning rate scheduler: cosine \\
& weight decay: $0.01$ \\
\cmidrule{1-2}
supervised+ppo & steps per update: $240$\\
    & total number of steps: $96,000$ \\
    & batch size: $24$ \\
    & epochs per update: $5$ \\
    & learning rate: $0.000001$ \\
    & entropy coefficient: $0.001$ \\
    & initial kl coeff: $0.00001$ \\
    & target kl: $3$ \\
    & discount factor: $0.99$ \\
    & gae lambda: $0.95$ \\
    & clip ratio: $0.2$ \\
    & value function coeff: $0.5$ \\
    & rollouts top k: $100$ \\
\cmidrule{1-2}
decoding & sampling: True \\
& temperature: $0.7$ \\
& min length: $5$ \\
& max new tokens: $20$\\
\cmidrule{1-2}
tokenizer & padding side: left\\
& truncation side: right\\
& max length: 512 \\
\bottomrule
\end{tabular}
}
\caption{List of hyperparameters for Alphabetization.}
       \label{tbl:alph_hyperparams}
\end{table}
\begin{table}
\centering
\footnotesize
\resizebox{0.5\textwidth}{!}{
\begin{tabular}{ll}
\toprule
\textbf{Model Params}
& \multicolumn{1}{c}{\textbf{value}} \\ 
\cmidrule{1-2}
supervised & batch size: $5$\\
& epochs: $5$ \\
& learning rate: $0.00001$ \\
& learning rate scheduler: cosine \\
& weight decay: $0.01$ \\
\cmidrule{1-2}
supervised+ppo & steps per update: $256$ \\
    & total number of epochs: $256,000$ \\
    & batch size: $8$ \\
    & epochs per update: $5$ \\
    & learning rate: $0.0000005$ \\
    & entropy coefficient: $0.001$ \\
    & initial kl coeff: $0.01$ \\
    & target kl: $2$ \\
    & discount factor: $0.99$ \\
    & gae lambda: $0.95$ \\
    & clip ratio: $0.2$ \\
    & value function coeff: $0.5$ \\
    & rollouts temperature: $0.7$ \\
\cmidrule{1-2}
decoding & sampling: True \\
& temperature: $0.3$ \\
& min length: $15$ \\
& max new tokens: $50$\\
& repetition penalty: $0.2$ \\
\cmidrule{1-2}
tokenizer & padding side: left\\
& truncation side: left\\
& max length: 512 \\
\bottomrule
\end{tabular}
}
\caption{List of hyperparameters for Interscript.}
       \label{tbl:plan_hyperparams}
\end{table}
\begin{table}
\centering
\footnotesize
\resizebox{0.5\textwidth}{!}{
\begin{tabular}{ll}
\toprule
\textbf{Model Params}
& \multicolumn{1}{c}{\textbf{value}} \\ 
\cmidrule{1-2}
supervised & batch size: $4$\\
& epochs: $7$ \\
& learning rate: $0.00001$ \\
& learning rate scheduler: cosine \\
& weight decay: $0.01$ \\
\cmidrule{1-2}
supervised+ppo & steps per update: $1024$ \\
    & total number of epochs: $143,360$ \\
    & batch size: $4$ \\
    & epochs per update: $3$ \\
    & learning rate: $0.0000001$ \\
    & entropy coefficient: $0.001$ \\
    & initial kl coeff: $0.01$ \\
    & target kl: $2$ \\
    & rollouts temperature: $0.7$ \\
\cmidrule{1-2}
decoding & sampling: True \\
& temperature: $0.7$ \\
& min length: $20$ \\
& max new tokens: $150$\\
& repetition penalty: $0.2$ \\
\cmidrule{1-2}
tokenizer & padding side: right\\
& truncation side: right\\
& max length: 1024 \\
\bottomrule
\end{tabular}
}
\caption{List of hyperparameters for Topic-Based Summarization dataset by \citet{saunders2022self}.}
       \label{tbl:summ_hyperparams}
\end{table}

\subsection{Results for \textit{Self-Refine} \cite{madaan2023self} on Alphabetization}
\label{app:self-refine}
\citet{madaan2023self} and \citet{chen2023teaching} propose that self-generated critiques (sampling critiques simply via few-shot prompting) is useful for a range of tasks. We examine if self-generated critiques are more useful for Alphabetization task than other techniques proposed in \cref{tb:alph}. Doing so, we curate a few-shot prompt:

{
\footnotesize
\begin{verbatim}

Below is a given list of words which are 
supposed to be sorted in alphabetical order.
Describe what is wrong in the provided 
ordering.

---

Ordering: quirky whimsical bubbly joyous 
delightful melodic glimmering vivacious 
radiant lively zestful spontaneous
Feedback: Whimsical should come in the end. 
Delightful should come before joyous.

---

Ordering: airy amiable animated ardent 
astute beaming blithe brilliant
Feedback: This listed is correctly sorted.

---

Ordering: curious sprightly vivacious 
tenacious passionate vivacious
Feedback: The list contains duplicates 
and passionate should come before 
sprightly.

---

Ordering: {ordering}
Feedback:
\end{verbatim}
}

After sampling critiques using the above prompt, we follow the same steps described in \cref{sec:exps} for refinement. We obtain 21.6\% exact match accuracy using these critiques which is a significant drop from \texttt{code-davinci-002}'s initial performance. When we sample critiques and refinements from \texttt{text-davinci-002}, the exact match score increases notably (to 58.6\%) while still hurting the initial accuracy of 63.6\%. Self-Refine may improve over initial performance when critiques are sampled from more capable models such as \texttt{text-davinci-003} \cite{madaan2023self}. Nonetheless, having initialized as T5-large, RL4F's critique model already produces useful feedback despite being significantly smaller than \texttt{text-davinci-003}.

\section{Sample Predictions}
Sample predictions for all three tasks are provided in \cref{tb:pred_samples_plan}, \cref{tb:pred_samples_alph} and \cref{tb:pred_samples_summ}. A manual examination of the generated critiques revealed that close to 100\% of the critiques generated for alphabetization and action planning are grammatical. However, critiques for the topic-based-summarization task often involve repetitions or generic calls to fix or improve the answer e.g. "The answer is not adequate. The answer is wrong, it is wrong and should be fixed."
\label{sec:sample_predictions}
\begin{table*}
\resizebox{\textwidth}{!}{%
\begin{tabular}{lp{4cm}p{4cm}p{4cm}p{4cm}}
\toprule
Input                                                                                                                                                                                                                                                                                                                                                                      & \supervised Critique                                                                                           & Edit w/ \supervised                     & RL4F Critique                                                                                         & Edit w/ RL4F                    \\ \midrule
\begin{tabular}[c]{@{}l@{}}Goal: go to the hardware store\\ Steps:\\ 1. pay for the items \\ 2. decided to go to the hardware store \\ 3. pack items in a bag \\ 4. create a shopping list \\ 5. go to the grocery store \\ 6. drive to the store \\ 7. walk around the store \\ 8. put items in cart \\ 9. pay for the groceries\end{tabular}                              & You can't pay for groceries if you don't pay for the groceries. {\textcolor{red}{[Repetition]}}                                              & Insert 'pay for the groceries' before 9  \textcolor{red}{[Repeated nodes due to critique.]}    & The plan tells you to pay for the items before you go to the hardware store.  \textcolor{blue}{[Correctly points out an error.]}                        & Reorder the edge between 1 and 2 \textcolor{blue}{[GPT-3 correctly identifies the node to be replaced,} \textcolor{red}{yet the new location is not correct.]}     \\ \midrule
\begin{tabular}[c]{@{}l@{}}Goal: ride the bus\\ Steps:\\ 1. decided to ride the bus\\ 2. put on shoes and clothing\\ 3. walk to the front door\\ 4. grab the handle\\ 5. open the door\\ 6. walk out of the house\\ 7. lock the door\\ 8. ride the bus\end{tabular}                                                                                                         & You have to be in the house before you can get out. You can't get out of the house until you're in the house. \textcolor{red}{[Logically sound, but unhelpful critique.]} & Reorder the edge between 5 and 6  \textcolor{red}{[GPT-3 fixes what it thinks the critique is saying.]}           & You have to be at the bus stop before you can ride it.  \textcolor{blue}{[Points out an helpful correction.]}                                              & Insert 'walk to the bus stop' after 8 \textcolor{blue}{[A missing node is added} \textcolor{red}{to an incorrect place.]}\\\midrule
\begin{tabular}[c]{@{}l@{}}Goal: turn on computer\\ Steps:\\ 1. sit down at computer\\ 2. pull the chair closer to the computer\\ 3. look for the power switch on the computer\\ 4. check the power connection\\ 5. move a finger to the power switch\\ 6. press the power switch on the computer\\ 7. relax until the computer boots up\\ 8. turn on computer\end{tabular} & You can't press the power switch on the computer if you don't have the power switch.     \textcolor{red}{[Unhelpful critique.]}                     & Insert node 'find the power switch' before 6 \textcolor{red}{[Critique results in incorrect edit.]} & You have to press the power switch to turn on the computer. You have to do this before you can relax. \textcolor{red}{[Points out what's already correct in the plan.]}& Reorder the edge between 6 and 7  \textcolor{red}{[Critique results in incorrect edit.]}   \\\midrule
\begin{tabular}[c]{@{}l@{}}Goal: pick up large kettle from counter\\Steps:\\1. make sure ingredients are all ready\\2. look for the pot\\3. extend hand out\\4. reach for the pot\\5. grip the pot\\6. lift the pot\\7. pick up pot from counter\end{tabular} & You need to reach for the pot before you can reach for it.     \textcolor{red}{[Incorrect critique involving repetition.]}                     & Reorder the edge between 3 and 4 \textcolor{red}{[Critique results in an unnecessary edit.]} & Plan tells you to reach for the kettle before you reach for it. \textcolor{blue}{[Points out a redundancy in the plan,]}& Remove node 3  \textcolor{blue}{[Removes unnecessary node.]}   \\
\bottomrule
\end{tabular}%
}
\caption{Samples for Interscript \cite{tandon2021interscript}. The authors' comments evaluating the generations are given in brackets. Red comments indicate what went wrong and blue means correct. Refinements use the corresponding critiques.}
\label{tb:pred_samples_plan}
\end{table*}
\begin{table*}[]
\resizebox{\textwidth}{!}{%
\begin{tabular}{p{4cm}p{4cm}p{4cm}p{4cm}p{4cm}}
\toprule
Input & \supervised Critique & \supervised Refinement & RL4F Critique  & RL4F Refinement \\ \midrule
accomplishments catwalk accessory incidental sixty gad remission melted juices fellowship unto dredge {\Large |||} accessory accomplishments catwalk dredge fellowship gad incidental juices melted remission sixty unto & The word melted is placed in an incorrect position. \textcolor{red}{[The word melted is in the correct position.]} & accessory accomplishments catwalk dredge fellowship gad incidental \textcolor{red}{melted juices} remission sixty unto & The list is correctly sorted. \textcolor{blue}{[Correct.]}& accessory accomplishments catwalk dredge fellowship gad incidental juices melted remission sixty unto \textcolor{blue}{[Correct.]}\\ \midrule
inseparable snapping astrology advertising unjust indies dismantle bangkok boosters brasil businessmen {\Large |||} advertising astrology bangkok brasil businessmen boosters dismantle indies inseparable injustice snapping & The word unjust is missing. \textcolor{blue}{[Correctly points out that unjust is missing.]} & advertising astrology bangkok brasil businessmen boosters dismantle indies injustice inseparable snapping \textcolor{red}{[GPT-3 fails to add unjust; keeps the word injustice which was not in the original list.]}& The word unjust is replaced with injustice. \textcolor{blue}{[Correctly indicates unjust is replaced with injustice]}& advertising astrology bangkok brasil businessmen boosters dismantle indies inseparable injustice snapping \textcolor{red}{[GPT-3 fails to replace injustice with unjust.]}\\
\bottomrule
\end{tabular}%
}
\caption{Samples for Alphabetization. Unsorted list and GPT-3's attempt in sorting them are separated with ||| in the input.}
\label{tb:pred_samples_alph}
\end{table*}
\begin{table*}[]
\resizebox{\textwidth}{!}{%
\begin{tabular}{p{4cm}p{4cm}p{4cm}p{4cm}p{4cm}}
\toprule
Input & \supervised Critique & \supervised Refinement & RL4F Critique  & RL4F Refinement \\ \midrule
Jeffery was sitting in his statistics class nodding away as the information was simply passing by him. He could hardly wait to get out of class. Mondays, and Fridays were fine because there was a small chance of him seeing her. However, on Wednesdays she was always on time. He could see it now walking out from his this room strolling down Taylor Mall.[\dots] & The response does not mention that Jeffery's friend is killed. The response only mentions that he is saved. \textcolor{red}{[No mention of murder in the story.]} & Summer is a student in Jeffery's statistics class. She is a beautiful girl with long dark hair and a sunflower dress. Every Wednesday at 12:30, she waits for Jeffery to come out of class so she can see his smiling face. However, she has never had the courage to ask him to a date. \textcolor{red}{[GPT-3 ignores the critique and leaves the answer as is. It's Jeffery who is interested in asking Summer out not the other way around.]}  & The answer doesn't mention Jeffery asking her to a date. It should mention that he is not sure how to ask her to a date. \textcolor{blue}{[This is an important detail that's is confused by the model's initial prediction.]} & Summer is a student in Jeffery's statistics class. She is a beautiful girl with long dark hair and a sunflower dress. Every Wednesday at 12:30, she waits for Jeffery to come out of class so she can see his smiling face. However, she has never had the courage to ask him to a date. Jeffery is not sure how to ask her to a date. \textcolor{blue}{[GPT-3 successfully adds that it's Jeffery who is into Summer, }\textcolor{red}{ yet it fails to remove the previous incorrect sentence suggesting otherwise.]} \\ \midrule
Gary left the house at six am on Wednesday; headed out to endure another long workday. After he walked through the yard to the driveway, he made sure to latch the gate securely, because that’s what Sarah liked. He smiled as he did so, and thought of how earlier, he had closed the shower curtain all the way (another Sarah preference), and gently lowered the toilet seat. [\dots]Summarize how Gary feels about his life.

Answer: Gary feels that his life is boring and that he is stuck in a rut. He feels that he is a slave to his wife and his job. He wants to be free of his responsibilities and wants to be a bartender. & The answer should mention that Gary gets a friend for dinner. There are other examples of a friendship that are mentioned in the text. \textcolor{red}{[The feedback is not helpful with respect to the question which asks how Gary is feeling about life.]} & Gary feels that his life is boring and that he is stuck in a rut. He feels that he is a slave to his wife and his job. He wants to be free of his responsibilities and wants to be a bartender. However, he gets a friend for dinner and there are other examples of a friendship that are mentioned in the text. \textcolor{red}{[GPT-3 appends the irrelevant details suggested by the critique.]}& It's omitted that he is the one who is angry. He has a son, Steven. \textcolor{blue}{[Critique suggests a detail about his anger.]} & Gary feels that his life is boring and that he is stuck in a rut. He feels that he is a slave to his wife and his job. He wants to be free of his responsibilities and wants to be a bartender. He is angry at his wife and has a son, Steven. \textcolor{blue}{[Having seen the critique finds out that Gary's anger is actually towards his wife and improves its answer accordingly.]} \\ \midrule
SR-25 pattern magazine

An SR-25, AR-10 or LR-308 PATTERN MAGAZINE is a type of detachable firearm magazine based on the original Armalite AR-10 “waffle” magazine design introduced in the late 1950s. [\dots]What does the text say about the SR-25, AR-10, or LR-308 Pattern Magazine?

Answer: A detachable firearm magazine based on the original Armalite AR-10 'waffle' magazine design introduced in the late 1950s, the design has been used on many AR-10 derivatives. It was reintroduced in 2012 with their new AR-10A model. & The response includes inaccurate information. Firstly, a reference to the SR-25 pattern is omitted. \textcolor{red}{[The response does not contain inaccurate information and AR10A indeed contains SR25 pattern.]} & A detachable firearm magazine based on the original Armalite AR-10 'waffle' magazine design introduced in the late 1950s, the design has been used on many AR-10 derivatives. It was reintroduced in 2012 with their new AR-10A model. \textcolor{red}{[GPT-3 repeats the initial answer.]} & The response is missing several important information about the magazine. The first sentence should be more specific. \textcolor{blue}{[Critique calls for a more detailed summary.]}& A detachable firearm magazine based on the original Armalite AR-10 'waffle' magazine design introduced in the late 1950s, the design has been used on many AR-10 derivatives. It was reintroduced in 2012 with their new AR-10A model. The magazine is used in many AR-10 derivatives such as the Knight's Armament SR-25 and DPMS Panther LR-308, as well as bolt action rifles such as the Mossberg MVP.308, Ruger Precision Rifle and Q FIX. Aftermarket magazines are produced by Magpul, Lancer Systems, and others. \textcolor{blue}{[More comprehensive summary of magazine designs.]} \\
\bottomrule
\end{tabular}%
}
\caption{Samples for Topic-Based Summarization by \citet{saunders2022self}. Passages are truncated.}
\label{tb:pred_samples_summ}
\end{table*}

\section{Learning Curves for Reinforcement Learning}
\begin{figure}[!]
     \centering
     \begin{subfigure}[b]{0.5\textwidth}
         \centering
         \includegraphics[width=\textwidth]{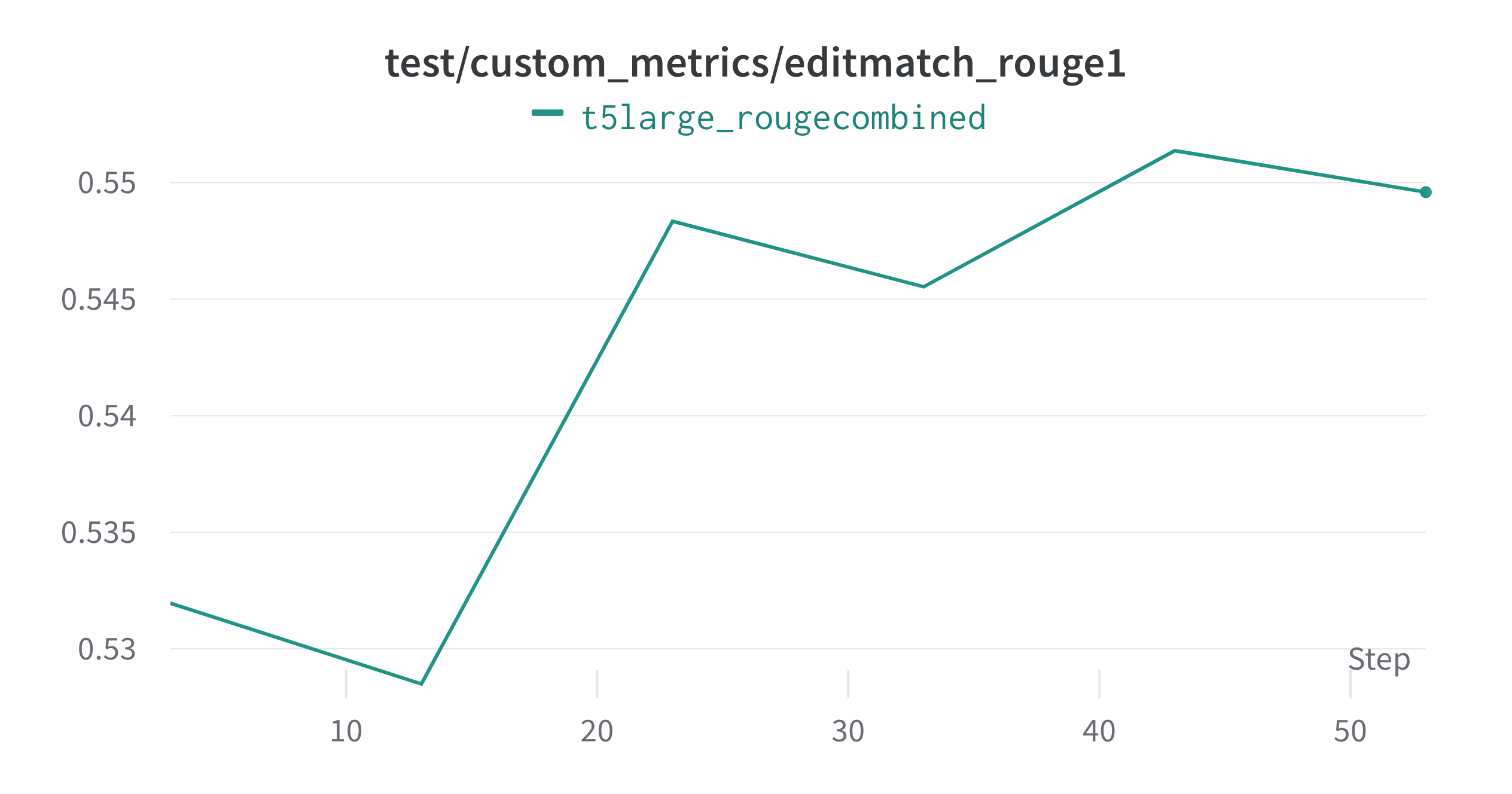}
         \caption{ROUGE for topic-based summarization.}
         \label{fig:lc_summ}
     \end{subfigure}
     \hfill
     \begin{subfigure}[b]{0.5\textwidth}
         \centering
         \includegraphics[width=\textwidth]{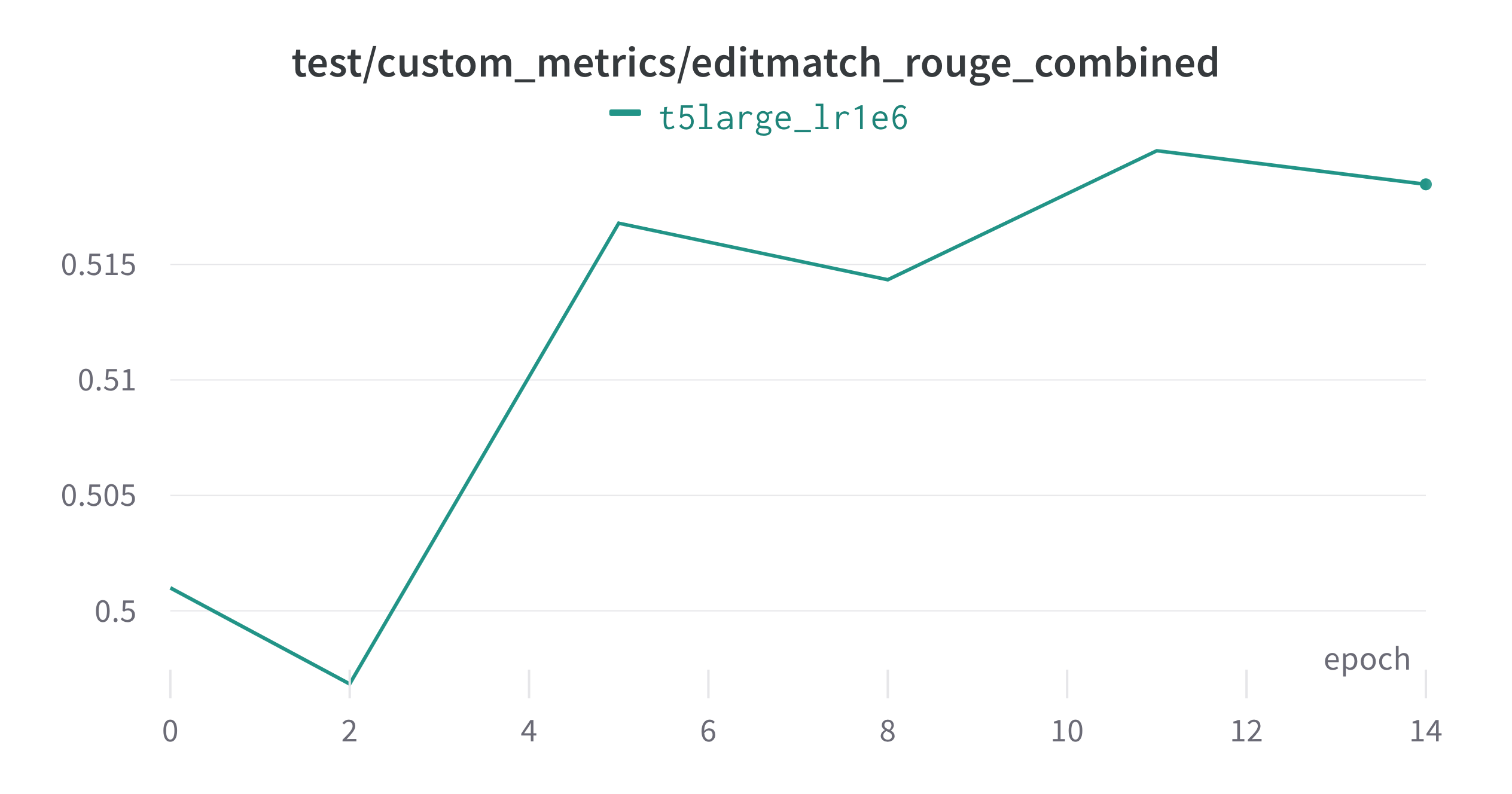}
         \caption{Mean ROUGE for Interscript.}
         \label{fig:lc_interscript}
     \end{subfigure}
     \vfill
     \begin{subfigure}[b]{0.5\textwidth}
         \centering
         \includegraphics[width=\textwidth]{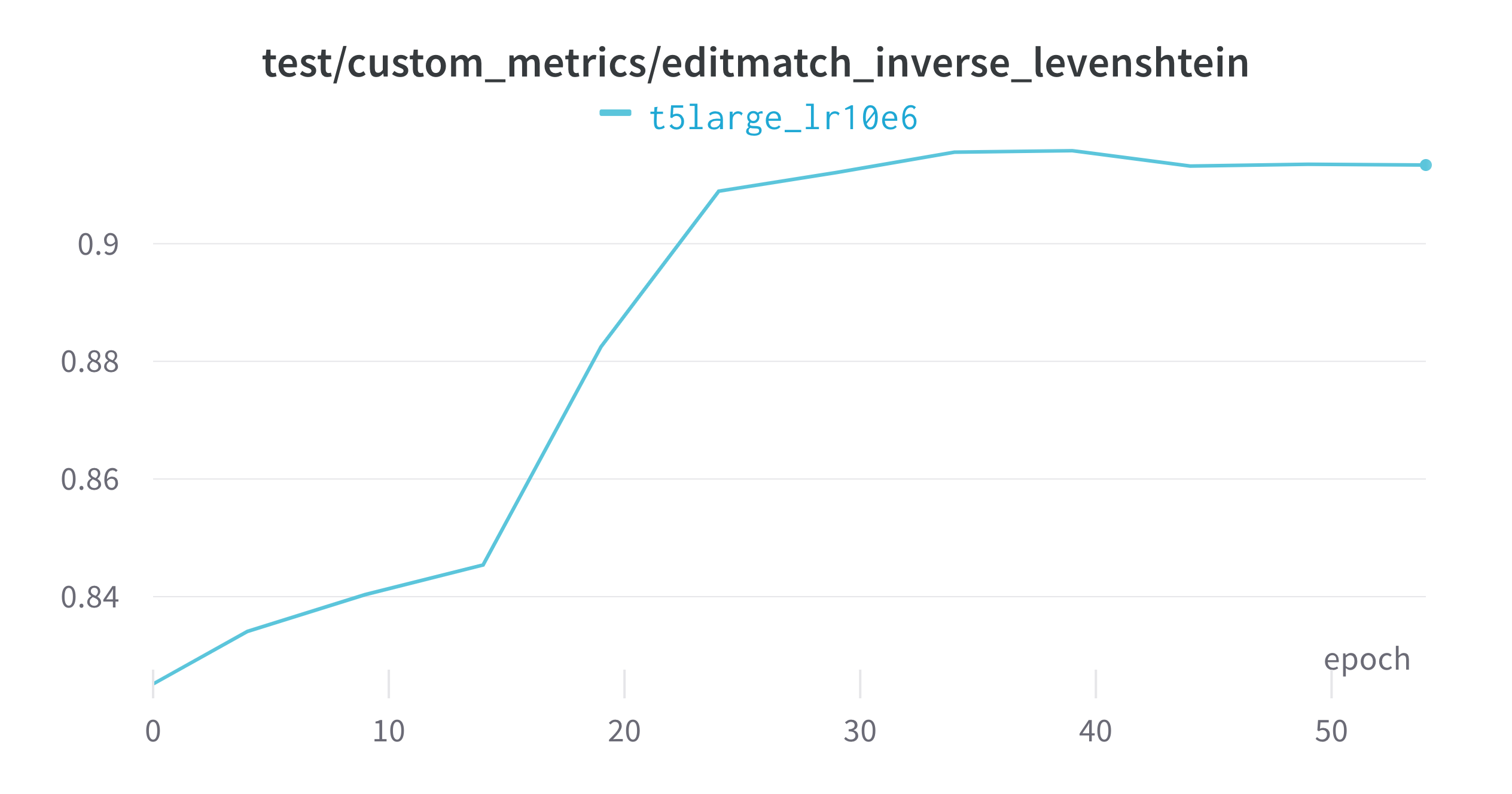}
         \caption{Inverse Levenstein distance for alphabetization.}
         \label{fig:lc_alpha}
     \end{subfigure}
     \hfill
     \begin{subfigure}[b]{0.5\textwidth}
         \centering
         \includegraphics[width=\textwidth]{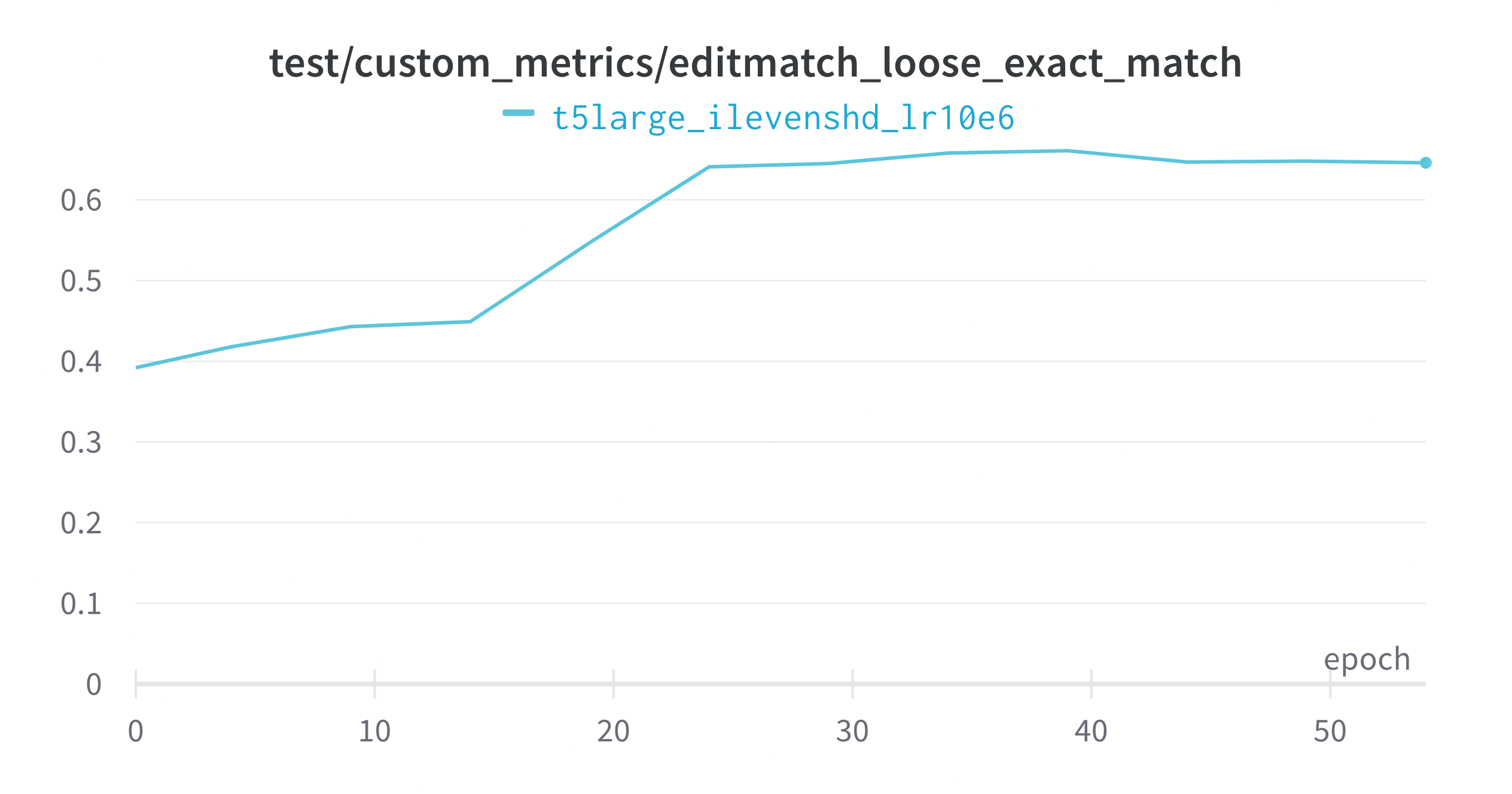}
         \caption{Exact match for alphabetization.}
         \label{fig:lc_alpha_em}
     \end{subfigure}
        \caption{As RL4F is trained, we track how evaluation metrics evolve for dev and test sets. Here, we display the results assessing the revised outputs conditioned on the critiques for test samples.}
        \label{fig:lc}
\end{figure}

In \cref{fig:lc}, we provide how evaluation metrics progress as 
\critic in RL4F is trained.




\end{document}